\documentclass[10pt,twocolumn,letterpaper]{article}

\usepackage{cvpr}              

\usepackage{algorithm}
\usepackage{algpseudocode}  
\usepackage{amsmath}
\usepackage{algorithm}
\usepackage{algpseudocode}
\usepackage{multicol}
\usepackage{lipsum}
\usepackage{xcolor}
\usepackage{colortbl}
\usepackage{graphicx}
\usepackage{array}
\usepackage{svg}
\usepackage{float}
\usepackage{booktabs}
\usepackage{multirow}
\usepackage{listings}
\usepackage{amsmath}
\usepackage[table]{xcolor}   

\algrenewcommand\algorithmiccomment[1]{\hfill\(\triangleright\) #1}

\definecolor{cvprblue}{rgb}{0.21,0.49,0.74}
\usepackage[pagebackref,breaklinks,colorlinks,allcolors=cvprblue]{hyperref}

\def\paperID{xxxxx} 
\def\confName{CVPR}
\def\confYear{2026}

\title{Latent Diffusion Inversion Requires Understanding the Latent Space}

\author{Mingxing Rao, Bowen Qu, Daniel Moyer\\
Vanderbilt University\\
Nashville, TN 37235,USA\\
{\tt\small \{mingxing.rao, bowen.qu, daniel.moyer\}@vanderbilt.edu}
}

\begin{document}
\maketitle
\begin{abstract}
The recovery of training data from generative models (``model inversion'') has been extensively studied for diffusion models in the data domain as a memorization/overfitting phenomenon. Latent diffusion models (LDMs), which operate on the latent codes from encoder/decoder pairs, have been robust to prior inversion methods. In this work we describe two key findings: (1) the diffusion model exhibits non-uniform memorization across latent codes, tending to overfit samples located in high-distortion regions of the decoder pullback metric; (2) even within a single latent code, memorization contributions are unequal across representation dimensions. Our proposed method to ranks latent dimensions by their contribution to the decoder pullback metric, which in turn identifies dimensions that contribute to memorization. For score-based membership inference, a sub-task of model inversion, we find that removing less-memorizing dimensions improves performance on all tested methods and datasets, with average AUROC gains of 1--4\% and substantial increases in TPR@1\%FPR (1--32\%) across diverse datasets including CIFAR-10, CelebA, ImageNet-1K, Pokémon, MS-COCO, and Flickr. Our results highlight the overlooked influence of the auto-encoder geometry on LDM memorization and provide a new perspective for analyzing privacy risks in diffusion-based generative models.

\end{abstract}

\begin{figure*}[t]
    \centering
    \includegraphics[width=\textwidth]{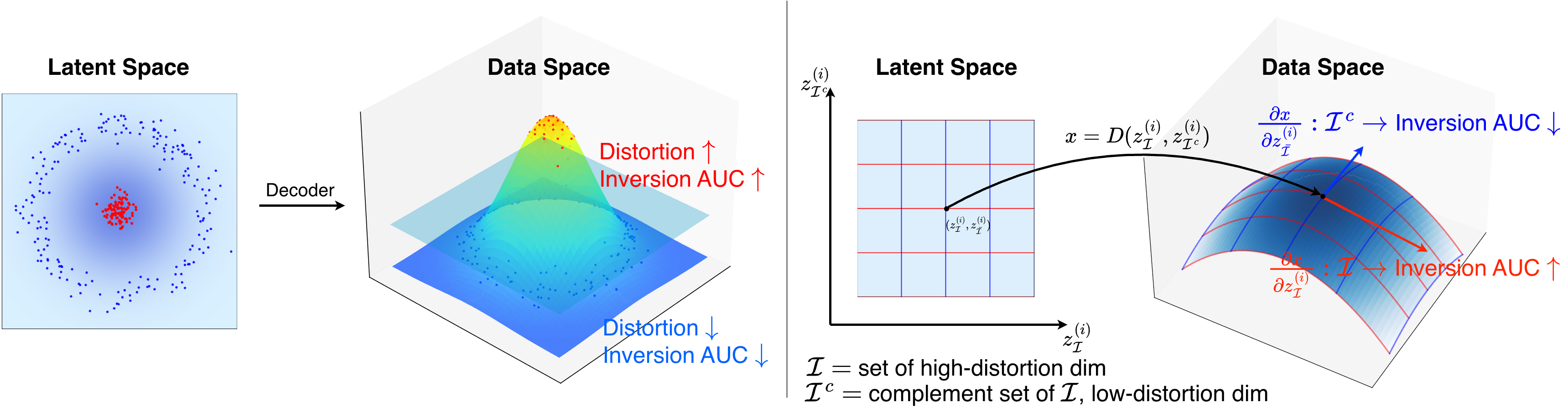}
    \caption{\textbf{Left:} Points mapped to high-distortion regions of the latent space (\textcolor{red}{red}) are more vulnerable to inversion and exhibit stronger memorization compared to those in low-distortion regions (\textcolor{blue}{blue}). \textbf{Right:} Latent grid lines and their decoded counterparts show that high-distortion latent dimensions (\textcolor{red}{red}) induce larger changes in data space, whereas low-distortion dimensions (\textcolor{blue}{blue}) induce smaller changes. This reflects that latent diffusion models memorize high-distortion dimensions more strongly than low-distortion ones.}

    \label{fig:mainfig}
\end{figure*}

\section{Introduction}
Model inversion is the task of recovering training samples from a trained generative model.
Generative models should ostensibly be fitting an underlying data distribution; recovering individual training points from their model weights implies an overfitting or memorization phenomenon.
Inversion methods may be informative of pathological structures and fitting behavior, both for fitted models and more broadly for the model family. In this paper we focus on Latent Diffusion Models (LDMs)~\cite{rombach2022high} and pathological overfitting behaviors resulting in their invertibility.

A crucial precursor to model inversion is the \emph{membership inference attack} (MIA), where attack methods attempt to distinguish training examples (members) from non-training examples (non-members). This measures how well we can identify training-points without considering the relatively unrelated search problem over the space of images. Many effective MIA methods have been proposed for diffusion models \cite{carlini2023extracting, zhai2024membership, kong2023efficient, matsumoto2023membership, fu2023probabilistic}. A model with stronger training-set memorization (overfit) typically shows higher MIA vulnerability; thus, the sensitivity to MIA serves as a practical metric for evaluating memorization.

The inversion of diffusion models has been extensively explored in prior work~\cite{zhu2016generative,creswell2018inverting,carlini2023extracting,somepalli2023diffusion,somepalli2023understanding,gu2023memorization}. However, for Latent Diffusion Models (LDMs)~\cite{rombach2022high}, existing research at once (1) predominantly focuses on inverting the diffusion process itself, treating the latent space as a fixed substrate and largely overlooking the role of the associated variational auto-encoder (VAE), and (2) has significantly reduced inversion performance \cite{rao2025score} relative to data-domain diffusion processes.

Prior work~\cite{rao2025score} shows that varying the latent-space regularization leads to substantially different degrees of vulnerability to membership inference, indicating that the structure of the latent space influences model memorization.
In this paper we take this a step further, demonstrating that geometric properties of the decoder implicate upon membership inference performance.
Based on statistics of the decoder geometry, we construct a filtering procedure that improves membership inference performance \emph{across all tested methods}; this demonstrates that these local statistics are connected with overfit phenomena, and not simply biases or idiosyncrasies in particular model inversion techniques.





To characterize geometric properties of the decoder, we employ the \textit{pullback metric} from Riemannian geometry~\cite{arvanitidis2017latent,chadebec2022geometric,arvanitidis2020geometrically}.
This is metric tensor associated with the decoder mapping (the ``pulling back'' of the data-space metric to the latent space),
and its determinant measures the local volume change induced by the decoder~\cite{arvanitidis2017latent}, which we call the distortion.
Empirically, we find that latent diffusion models tend to memorize samples located in regions of \emph{high} decoder distortion more strongly than those in low-distortion regions (diagrammed in Fig. \ref{fig:mainfig}).
In contrast, if the latent space exhibits nearly uniform distortion, memorization becomes correspondingly uniform.
The strong correlation between decoder's local distortion and memorization indicates that the choice of encoder/decoder architecture plays an important role in score field learning. Experiments to support these claims are presented in Section~\ref{exp:local_dist}.

Previous work observed that data domain diffusion denoisers preferentially reconstruct high-variance directions while oversmoothing low-variance directions \cite{lukoianov2025locality,wang2023hidden,kadkhodaie2023generalization}. These directions correspond to eigenvectors of the data covariance matrix. Although this analogy is not rigorously transported to the latent space without significant computation, whose nonlinear encoding behavior can differ substantially point-to-point, it suggests that certain latent directions may contribute disproportionately to memorization. We improve the inversion performance by preferentially including or excluding different latent space dimensions based on the \emph{per-dimensional contribution} to the local distortion induced by the decoder. In Section~\ref{subsec:per-dimension_influence}, we introduce a computationally efficient method to quantify these contributions, diagrammed in Fig. \ref{fig:mainfig} (right). We empirically validate this idea by selectively removing the less-contributing (alternatively, less-memorizing) dimensions and evaluating the resulting membership inference attacks. Across multiple datasets, this dimensional masking consistently improves attack performance of all attack methods: AUROC increases by 1--4\%, and TPR@1\%FPR improves by 1--32\%. 

In summary, our contributions are the following:
\begin{enumerate}
    \item We show that latent diffusion models exhibit \emph{spatially non-uniform memorization} in latent space and demonstrate that this behavior is tightly correlated with the \emph{local distortion} measured by the decoder pullback metric.
    \item We propose a principled, geometry-driven measure of \emph{per-dimensional memorization strength} based on each latent dimension's contribution to decoder distortion.
    \item We provide extensive empirical evidence showing that removing less-memorizing latent dimensions of attack vector before computing the norm statistics consistently improves all metrics over all MIA methods.
\end{enumerate}
All data splits, model checkpoints, training/fine-tuning scripts, and testing code are released on our GitHub repository \url{https://github.com/mx-ethan-rao/VAE2Diffusion.git}.

\label{sec:intro}

\section{Related Work}

\paragraph{Riemannian Geometry of Variational Autoencoders:}
A line of work studies the Riemannian geometry induced by deep generative models, and VAEs in particular, via the pullback metric of the decoder. \citet{arvanitidis2017latent} show that the generator of a VAE induces a stochastic Riemannian metric on the latent space and that geodesics and distances computed under this metric lead to more semantically meaningful interpolations, improved sampling, and better clustering~\cite{arvanitidis2017latent,arvanitidis2020geometrically}. \citet{shao2018riemannian} further develop algorithms for computing geodesics and parallel transport on the manifolds defined by deep generative models. Subsequent work has used this geometric viewpoint to regularize latent spaces, design geometry-aware VAEs, and perform data augmentation in small-sample regimes by exploiting the learned Riemannian structure~\cite{chadebec2022geometric,arvanitidis2020geometrically}. These works primarily use the pullback metric to study interpolation, representation quality, and manifold geometry, while instead we study the decoder-induced pullback metric in terms of its \emph{privacy-relevant} changes to the original space. We estimate local distortion for high-dimensional VAEs used in LDMs, and we show that this distortion reveals where the latent diffusion model memorizes most strongly and which latent dimensions contribute most to membership leakage.

\paragraph{Latent-domain diffusion models:}
LDMs~\cite{rombach2022high} factorize generative modeling into a VAE and a diffusion model trained in the resulting latent space, dramatically reducing computational cost while preserving visual fidelity. Subsequent work has largely followed this paradigm but focused on improving synthesis quality, efficiency, or conditioning, e.g., by replacing the standard VAE with alternative autoencoder families~\cite{zheng2025diffusion,shi2025latent}, or by scaling latent diffusion to new modalities and tasks. In parallel, a growing literature analyzes privacy and memorization in diffusion models, proposing increasingly strong membership inference and data extraction attacks~\cite{carlini2023extracting,duan2023diffusion,kong2023efficient,fu2023probabilistic,zhai2024membership,pang2023white}; several of these membership inference studies show ``pass-through'' phenomena of data-domain denoisers~\cite{kadkhodaie2023generalization,li2024unveiling,lian2025unveiling}. However, these analyses either operate in pixel space or treat the autoencoder in LDMs as a fixed pre-processor. In contrast, we explicitly attribute memorization behavior of latent-space diffusion models to the \emph{geometry} of the encoder/decoder pair. Our results show that decoder-induced local distortion in the VAE latent space controls spatially non-uniform memorization and that per-dimension contributions to this distortion identify latent directions that dominate membership leakage, bringing attention to the encoder/decoder architecture as a primary factor in the inversion properties of LDMs.

Prior work~\cite{lukoianov2025locality,wang2023hidden} has shown that pixel-domain diffusion model denoisers often struggle with high-frequency components, which correspond to low-variance directions in the eigen-space of the data covariance. In particular, these denoisers tend to suppress high-frequency content and implicitly project reconstructions toward high-variance, low-frequency directions. However, in latent-domain diffusion models the diffusion process operates on nonlinear, semantically structured latent representations produced by an encoder, and the mapping back to the image domain is mediated by a decoder with nontrivial geometry. We empirically show that the local distortion of an image in latent space is not fully explained by its high-frequency energy in the pixel domain (see Table \ref{tab:correlation_dist_hf}). This suggests that frequency-based analyses developed for data-domain diffusion models~\cite{li2024unveiling,lian2025unveiling} may not directly transfer to latent-domain models, though there is a analogous non-linear effect.

Moreover, \citet{lian2025unveiling} report that suppressing high-frequency components in the denoised images can enhance the separation between membership attack statistics on training versus non-training data. In our experiments with Stable Diffusion, we do \emph{not} observe a similar improvement when applying analogous high-frequency suppression (see Appendix, Section~\ref{sec:attack_suppress_high_freq}). Taken together, these findings indicate that latent-domain diffusion models require a different analytical perspective: rather than relying solely on frequency-based reasoning in pixel space, one must account for the nonlinear encoder--decoder geometry and its induced distortion in latent space.

\label{sec:related_work}

\begin{figure}[t]
    \includegraphics[width=\columnwidth]{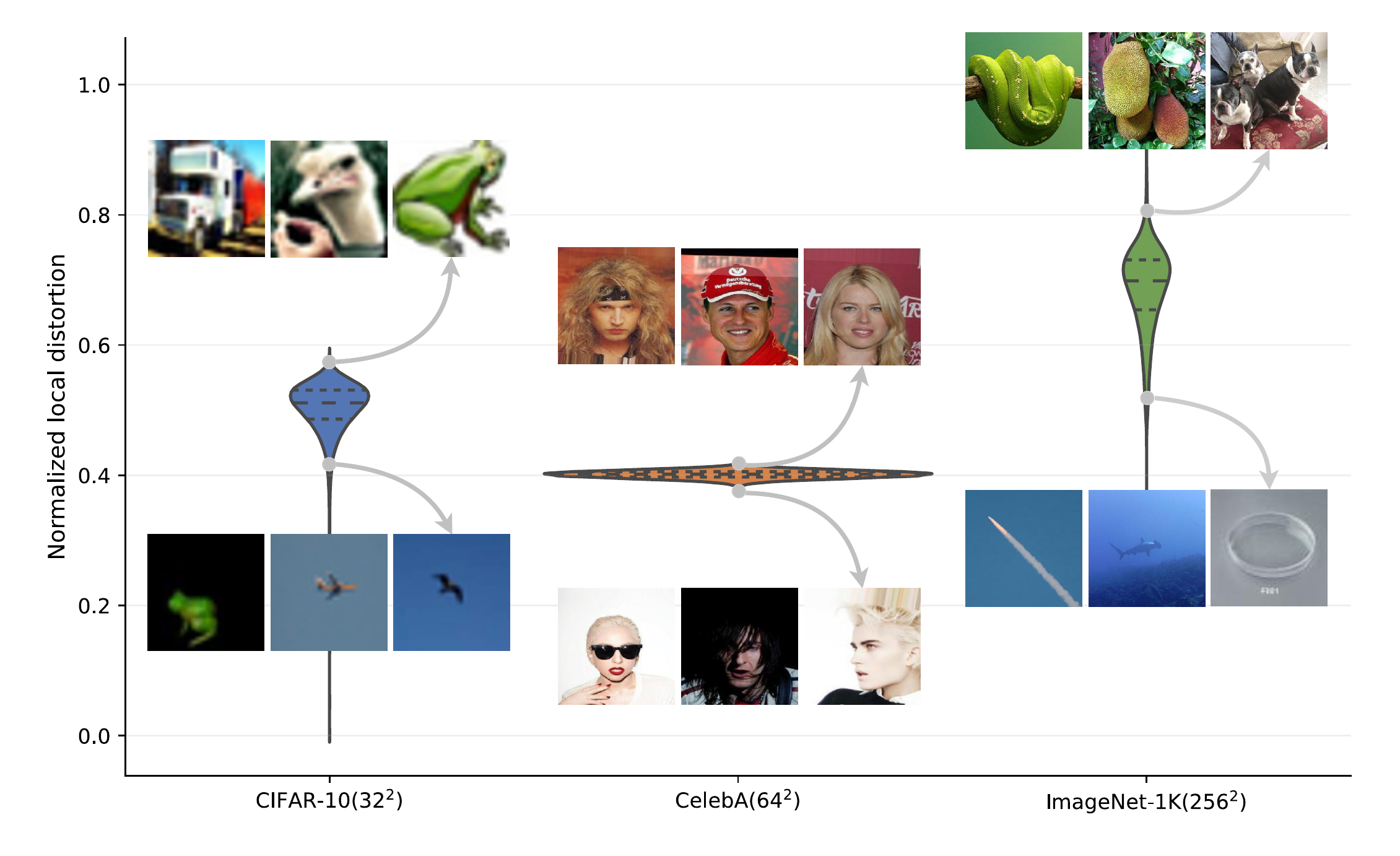}
    \caption{Each violin plot shows the distribution of the decoder-induced local distortion, with representative samples displayed at high-distortion (top) and low-distortion (bottom) regions. CelebA shows an almost uniform distortion distribution, while CIFAR-10 and ImageNet-1K display larger variation across samples. Refer to the appendix~\ref{sec:examples_local_distorch} for additional samples.   
    }
    \label{fig:distortion}
\end{figure}

\begin{figure*}[t]
    \centering
    \includegraphics[width=\textwidth]{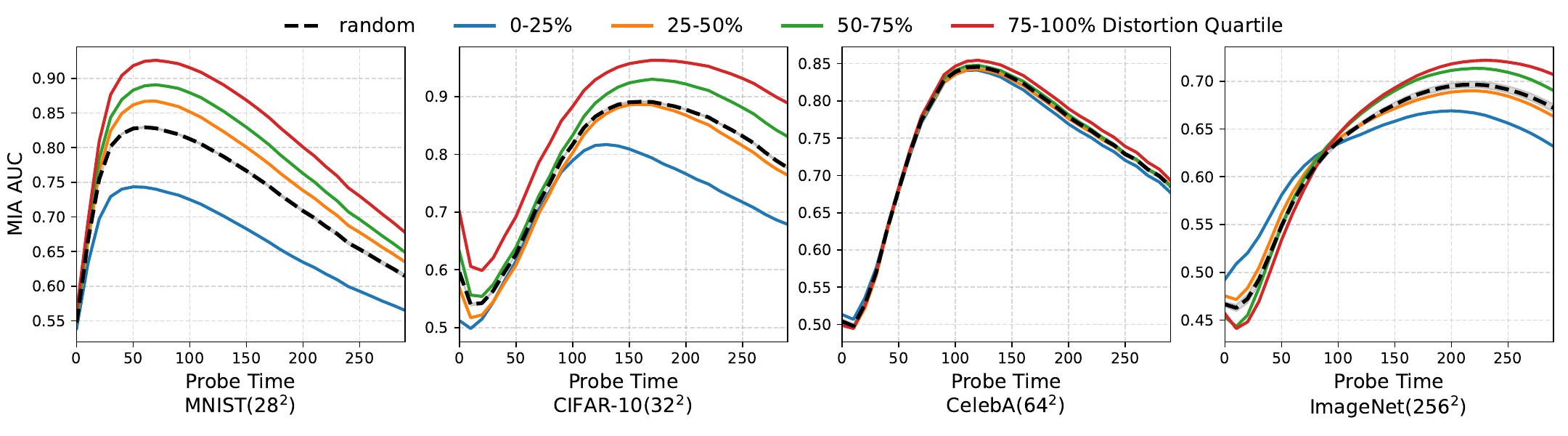}
    \caption{Membership inference AUC (measured by SimA) at different times for four datasets, stratified by quartiles of local distortion of decoder (0--25\%, 25--50\%, 50--75\%, 75--100\%). Quartile thresholds are computed jointly over members and held-out samples. Attacks are evaluated separately within each quartile, and a random baseline (mean and variance over ten trials) is shown for comparison. Higher-distortion quartiles consistently yield higher attack AUC.}

    \label{fig:mia_distort}
\end{figure*}

\section{Methodology}
\label{sec:method}

\textbf{Terminology:} Unless otherwise specified, the term \textit{latent space} refers to the latent representation learned by the VAE component of LDM. The diffusion process operating in this latent representation is referred to as the \textit{latent space diffusion model}.

We outline the conceptual development of our approach here. Section~\ref{subsec:score_mia} summarizes the state-of-the-art score-based membership inference attacker used to evaluate model memorization. Section~\ref{subsec:pullback_metric} introduces the definition of \textit{local distortion}, a key geometric property of the VAE decoder that quantifies how infinitesimal regions in latent space are stretched in the data space. In Fig.~\ref{fig:mia_distort} of Section~\ref{exp:local_dist}, we demonstrate that samples located in regions of higher local distortion are more susceptible to membership attacks, revealing a non-uniform latent space memorization of the latent space diffusion model. 

Motivated by this, we hypothesize that beyond non-uniform memorization across data samples, the latent space diffusion model exhibits \textit{dimension-wise non-uniform memorization}—that is, high-distorted dimensions tend to leak more membership information. Section~\ref{subsec:per-dimension_influence} introduces a simple method to quantify the influence of each dimension on memorization, and Section~\ref{sec:experiments} presents empirical evidence supporting this hypothesis.

\subsection{Score-based Membership Inference Attacks} \label{subsec:score_mia}

Membership Inference Attacks (MIA) are pre-cursor tasks to model inversion. They measure the separability (conditional on an attack methodology/model) of seen training and unseen test data given a trained diffusion model and a threat model.
While not performing the full search required by model inversion, MIA demonstrates that candidate data-points can be classified as \emph{in training} or \emph{out of training}. 
For our threat model, we consider a common grey-box \cite{duan2023diffusion, zhai2024membership, kong2023efficient, matsumoto2023membership, carlini2023extracting, fu2023probabilistic} scenario, where the attack can manipulate model inputs but should be agnostic to particular architectures or weight configurations.

The success of these attacks can be viewed as a proxy measurement for overfitting and memorization of the training set. While not exactly equivalent, at a high level their success indicates that the generative model being attacked has information about the training dataset beyond its underlying generative distribution, i.e., about the specific training dataset. Many of the attacks use this overfitting intuition to construct their methods, using overfit behaviors to identify training data.

A baseline “loss” attack (also referred to ``naive'' method) evaluates the denoising objective on the queried point by sampling $\varepsilon\!\sim\!\mathcal N(0,I)$, forming $x_t^\star=\sqrt{\bar\alpha_t}\,x^\star+\sqrt{1-\bar\alpha_t}\,\varepsilon$, and scoring
$$
\textsc{Loss}(x^\star,t)\;=\;\bigl\|\varepsilon-\hat\varepsilon_\theta(x_t^\star,t)\bigr\| ,
$$
often averaged over several $\varepsilon$ draws~\citep{matsumoto2023membership}.
SecMI further exploits diffusion structure by measuring a single–step \emph{posterior estimation error} (``$t$-error'') at a chosen timestep $t$ using deterministic DDIM mappings. 
Let $\Phi_\theta$ and $\Psi_\theta$ denote the deterministic reverse/denoise operators; with $\tilde x_t=\Phi_\theta(x^\star,t)$, SecMI’s statistic is
\[
\text{SecMI}(x^\star,t);=\;\bigl\|\Psi_\theta\!\bigl(\Phi_\theta(\tilde x_t,t),t\bigr)-\tilde x_t\bigr\|_2 ,
\]
which is thresholded (or fed to a small attack network) to decide membership~\citep{duan2023diffusion}.
PIA attains similar accuracy with far fewer queries by using the model’s own $t{=}0$ denoiser output as a \emph{proximal initialization} and then computing a forward/backward consistency error built from this initialization~\citep{kong2023efficient}. For $\tilde{x}_t^\star=\sqrt{\bar\alpha_t}x^* + \sqrt{1-\bar\alpha_t} \hat\varepsilon_{\theta}(x,t=0)$,
$$
\text{PIA}(x^\star,t) = \|\hat\varepsilon_{\theta}(x^\star,t=0) - \hat\varepsilon_{\theta}(\tilde{x}_t^\star,t)\| 
$$
Finally, SimA \cite{rao2025score} is the scaled Loss criteria at $\varepsilon=0$ (point estimate), which is the mean and mode of $\varepsilon$ distribution. This is inherently a direct score estimation.
\[
\textsc{SimA}(x^\star,t)\;=\;\bigl\|\hat\varepsilon_\theta(x^\star,t)\bigr\| ,
\]
motivated by theory linking the expected denoiser to a kernel–weighted local mean of nearby training samples. 
Empirically, SimA matches or exceeds multi–query baselines while being simpler and faster; hence we adopt SimA to quantify the model memoriztion in most experiments~\citep{rao2025score}.

\subsection{VAE Decoder Pullback Metric} \label{subsec:pullback_metric}


In a Variational Autoencoders (VAE), for many common architectures the decoder defines a smooth mapping 
\[
D: \mathcal{Z} \rightarrow \mathcal{X},
\]
that transforms a latent variable $z \in \mathcal{Z}$ into a data point $x = D(z) \in \mathcal{X}$. 
While $\mathcal{Z}$ is usually assumed to have Euclidean structure, this assumption is also usually incorrect when viewed in terms of the decoder and $\mathcal{X}$: the nonlinear decoder distorts distances, making Euclidean geometry inapplicable. 
Instead, we need to choose a more relevant metric for the space; one promising candidate from the literature is the decoder induced Riemannian metric on $\mathcal{Z}$, known as the \textit{pullback metric} \cite{arvanitidis2017latent, arvanitidis2020geometrically, shao2018riemannian}.

We write the Jacobian of the decoder at point $z$ as $J_D(z) = \frac{\partial D(z)}{\partial z}$.
The pullback metric is then
\begin{equation}
    G(z) = J_D(z)^{\top} J_D(z),
    \label{eq:pullback_metric}
\end{equation}
which is a symmetric positive semi-definite tensor describing how local directions in $\mathcal{Z}$ are stretched or compressed when mapped to $\mathcal{X}$. 
For an infinitesimal displacement $\mathrm{d}z$, the induced distance in data space is
\begin{equation}
    \lVert \mathrm{d}x \rVert^2 = \mathrm{d}z^{\top} G(z) \, \mathrm{d}z.
    \label{eq:riemannian_metric}
\end{equation}
A scalar measure of local geometric distortion is given by the volume change factor
\begin{equation}
    \mathrm{Distortion}(z)
    = \sqrt{\det(G(z))}.
    \label{eq:decoder_distortion}
\end{equation}
To compute this stably, we use its logarithmic form
\begin{equation}
    \log \sqrt{\det(G(z))} 
    = \sum_i \log \sigma_i(J_D(z)),
\end{equation}
where $\sigma_i(J_D(z))$ denote the singular values of the decoder Jacobian $J_D(z)$. 
Since computing all of the singular values of the Jacobian is intractable for high-dimensional latent spaces, we approximate the local distortion via a truncated spectral estimate, employing a \textit{Matrix-free Randomized Singular Value Decomposition (SVD)} to extract the top-$K$ singular values of $J_D(z)$ (see the pseudo-code in Appendix \ref{sec:approx_local_distor}). Specifically, $S_K=\sum_{i=1}^{K} \log \sigma_i(J_D(z)),$ for descending sorted $\sigma_i$. 

Large singular values appear to correspond to directions that produce fine-grained, spatially detailed (high-frequency) pixel changes.
If one image’s $S_K$ is much larger, Fig. \ref{fig:distortion} shows this image is usually semantically meaningful, which we assume that image’s local neighborhood contains more high-frequency detail (edges or textures) while the other image’s neighborhood is smoother or simpler. More examples and additional exposition on this phenonmenon in VAE can be found in Appendix \ref{sec:examples_local_distorch}. 

\subsection{Quantifying Per-dimension Influence on Latent Memorization}
\label{subsec:per-dimension_influence}

\paragraph{Influence Measure Definition:}
As we show empirically in Fig.~\ref{fig:mia_distort} of Section~\ref{exp:local_dist}, distortion is related to overfit phenomena.
We hypothesize that local to a given datapoint, latent dimensions contribute in differing amounts to these overfit phenomena.
Some latent dimensions contribute more strongly to the decoder’s local distortion are more susceptible to memorization.
To quantify this relationship, we define a per-dimension influence measure that captures each coordinate’s relative contribution to the VAE decoder’s Jacobian magnitude:
\begin{equation}
\label{eq:perdim-def}
\mathrm{Infl}_i(z) := \left\| \frac{\partial x}{\partial z_i} \right\|_2^2 = e_i^{\top} G(z)\, e_i = G_{ii}(z).
\end{equation}

\paragraph{Masking scheme for attack statistics:}
Given any scalar attack statistic, a \textbf{general paradigm} is $\|S(z)\|$, where $S(z)$ is an attack vector (e.g., score-loss, noise-prediction difference or loss aggregated across timestep) in latent space. We form a \emph{dimension-masked} statistic by keeping only a subset $\mathcal{I}\subset[d]$, where $d$ is the number of dimensions on latent space:
\begin{equation}
\label{eq:masked-stat}
\|S_{\mathcal{I}}(z)\|:= \|S( z) \odot \mathbb{I}_{\mathcal{I}}(i)\|, ~~\mathbb{I}_{\mathcal{I}}(i) =
\begin{cases}
1, & i \in \mathcal{I}, \\[4pt]
0, & i \notin \mathcal{I}.
\end{cases}
\end{equation}
We choose $\mathcal{I}$ as the top-$k$\% coordinates ranked by ${\mathrm{Infl}}_i$. Masking out low-influence coordinates removes directions that contribute
little but add noise to attack statistics. Empirically, this method \emph{increases} AUC and markedly boosts TPR@1\%FPR across datasets. The results are on Table \ref{tab:ldm_filtered} and \ref{tab:sd_filtered}. Noticeably, the proposed method depends on VAE only, which indicates that latent diffusion inversion requires understanding the latent space of VAE.


\noindent \textbf{Information allocation via the decoder geometry:}
The pullback metric $G(z)$ determines the local volume element $\sqrt{\det G(z)}$ and acts as a data-dependent
conditioning matrix on the training signal that the latent UNet receives through the reconstruction path.
Coordinates with larger $G_{ii}$ correspond to directions where small latent changes expand more in pixel space,
thus receiving \emph{higher effective signal-to-noise} during training and being more likely to carry sample-specific information.
Conversely, coordinates with tiny $G_{ii}$ transmit weak, noisy supervision and dilute $S(z)$ when included.

\noindent\textbf{Hutchinson Estimator: }Instead of explicitly computing the diagonal entries of the pullback metric $G_{ii}(z)=\|J_D(z)e_i\|_2^2$, which requires constructing the full decoder Jacobian $J_D(z)$, we estimate them efficiently using the \emph{Hutchinson stochastic trace estimator}~\cite{grathwohl2018ffjord}.
For any function $D:\mathcal{Z}\!\to\!\mathcal{X}$ with Jacobian $J_D(z)$, the diagonal of the Gram matrix can be expressed as
\begin{equation}
\operatorname{diag}(J_D^{\top}J_D)
\;=\;
\mathbb{E}_{v\sim\mathcal{N}(0,I)}
\bigl[
(J_D^{\top}v)\odot(J_D^{\top}v)
\bigr],
\end{equation}
where $\odot$ denotes the elementwise product.
In practice, this expectation is approximated by Monte Carlo averaging over $n_{\mathrm{mc}}$ random vectors $v$.
For each latent code $z$, we compute $J_D^{\top}v$ implicitly using reverse–mode autodifferentiation, avoiding explicit Jacobian construction.
The per–dimension influence is then defined as
\begin{equation}
\mathrm{Infl}_i(z)
\;=\;
\tfrac{1}{2}\log
\!\bigl(
\mathbb{E}_{v}
\bigl[(J_D^{\top}v)_i^{2}\bigr]
+\varepsilon
\bigr),
\end{equation}
where $\varepsilon$ is a small constant for numerical stability.

This estimator takes $\mathcal{O}(n_{\mathrm{mc}}\, T_{D})$, where $T_{D}$ denotes the computational cost of a single evaluation of the decoder $D$. It scales linearly in the number of Monte Carlo probes, and is independent of the explicit Jacobian size. Empirically, we found $n_{\mathrm{mc}}{=}8$ is good enough for image datasets of all resolutions. A pseudocode is provided in appendix ~\ref{sec:hutchinson_est}.

\begin{table*}[!t]
\centering
\caption{Attack performance of four baseline methods used to evaluate memorization of \textbf{LDMs}. The \textit{filtered} setting removes the top 40\% least-memorizing latent dimensions of the attack vector (as ranked by our method) before computing the norm. The first row provides an ablation in which 40\% of dimensions are \textit{randomly} removed. \textbf{Across all datasets and attack methods, the proposed \textit{filtered} setting improves all metrics.}}
\setlength{\tabcolsep}{5pt}
\renewcommand{\arraystretch}{1.05}
\scriptsize
\begin{tabular}{lccc|ccc|ccc}
\toprule
\multirow{2}{*}{Method} &
\multicolumn{3}{c|}{CIFAR-10 (\%)} &
\multicolumn{3}{c|}{CelebA (\%)} &
\multicolumn{3}{c}{ImageNet (\%)} \\
\cmidrule(lr){2-4}\cmidrule(lr){5-7}\cmidrule(lr){8-10}
& AUC↑ & ASR↑ & TPR@1\%FPR↑
& AUC↑ & ASR↑ & TPR@1\%FPR↑
& AUC↑ & ASR↑ & TPR@1\%FPR↑ \\
\midrule
SimA (random drop) & 86.44 & 78.80 & 15.92 & 83.05 & 75.69 & 10.20 & 69.30 & 64.90 &  4.16 \\
SimA               & 89.10 & 81.63 & 19.88 & 84.66 & 77.04 & 11.09 & 69.62 & 64.92 &  3.87 \\
\rowcolor{gray!20}
SimA (filtered)     & 91.26 & 83.58 & 24.56 & 88.18 & 80.25 & 17.03 & 72.55 & 67.01 &  7.77 \\
Loss               & 73.28 & 67.48 &  7.86 & 68.65 & 63.47 &  6.39 & 67.49 & 63.16 &  4.09 \\
\rowcolor{gray!20}
Loss (filtered)     & 75.87 & 69.81 &  9.69 & 70.12 & 64.68 &  7.16 & 69.99 & 64.68 &  7.01 \\
SecMI              & 87.42 & 80.12 & 19.11 & 83.09 & 75.85 & 10.53 & 68.21 & 63.44 &  3.71 \\
\rowcolor{gray!20}
SecMI (filtered)    & 89.93 & 81.97 & 23.84 & 86.74 & 78.92 & 16.47 & 71.02 & 65.88 &  7.49 \\
PIA                & 85.73 & 78.19 & 15.90 & 83.36 & 75.87 &  9.41 & 66.09 & 62.52 &  2.95 \\
\rowcolor{gray!20}
PIA (filtered)      & 88.49 & 80.65 & 20.08 & 87.41 & 79.28 & 21.13 & 70.74 & 65.58 &  4.75 \\
\midrule
\rowcolor{gray!50}
Mean $\Delta$ (\%)  & +2.51 & +2.15 & +3.86 & +3.17 & +2.73 & +6.09 & +3.22 & +2.28 & +3.10 \\
\rowcolor{gray!50}
Min $\Delta$ (\%)  & +2.16 & +1.85 & +1.83  & +1.47 & +1.21 & +0.77  & +2.50 & +1.52 & +1.80 \\

\bottomrule
\end{tabular}
\label{tab:ldm_filtered}
\vspace{-6pt}
\end{table*}

\begin{table*}[!t]
\centering
\caption{Attack performance of four baseline methods used to evaluate memorization of \textbf{Stable Diffusion}. The \textit{filtered} setting removes the top 40\% least-memorizing latent dimensions of the attack vector (as ranked by our method) before computing the norm. The first row provides an ablation in which 40\% of dimensions are \textit{randomly} removed. \textbf{Across all datasets and attack methods, the proposed \textit{filtered} setting improves all metrics.}}
\setlength{\tabcolsep}{5pt}
\renewcommand{\arraystretch}{1.05}
\scriptsize
\begin{tabular}{lccc|ccc|ccc}
\toprule
\multirow{2}{*}{Method} &
\multicolumn{3}{c|}{Pokémon (\%)} &
\multicolumn{3}{c|}{MS-COCO (\%)} &
\multicolumn{3}{c}{Flickr (\%)} \\
\cmidrule(lr){2-4}\cmidrule(lr){5-7}\cmidrule(lr){8-10}
& AUC↑ & ASR↑ & TPR@1\%FPR↑
& AUC↑ & ASR↑ & TPR@1\%FPR↑
& AUC↑ & ASR↑ & TPR@1\%FPR↑ \\
\midrule
SimA (random drop)  &  93.42 & 87.63 & 17.99 &  93.57 & 87.12 & 28.12 &  69.85 & 65.66 & 2.79 \\
SimA                & 93.50 & 87.87 & 20.38 & 93.71 & 87.34 & 29.80 & 70.04 & 65.96 &  2.59 \\
\rowcolor{gray!20}
SimA (filtered)      & 94.83 & 89.68 & 52.04 & 96.86 & 92.10 & 49.44 & 73.61 & 68.45 &  3.49 \\
Loss                & 92.03 & 85.47 & 40.77 & 83.26 & 75.80 & 13.48 & 63.13 & 59.74 &  2.35 \\
\rowcolor{gray!20}
Loss (filtered)      & 92.67 & 86.07 & 46.28 & 85.78 & 78.18 & 18.68 & 64.57 & 60.49 &  2.81 \\
SecMI               & 88.12 & 81.04 & 29.02 & 91.35 & 84.00 & 39.16 & 71.81 & 66.54 &  5.29 \\
\rowcolor{gray!20}
SecMI (filtered)     & 89.37 & 82.47 & 30.94 & 93.77 & 87.46 & 50.40 & 74.16 & 68.51 &  6.46 \\
PIA                 & 94.93 & 90.52 & 32.85 & 92.51 & 85.72 & 24.76 & 68.88 & 64.61 &  2.67 \\
\rowcolor{gray!20}
PIA (filtered)       & 96.23 & 91.47 & 53.00 & 96.03 & 90.30 & 43.24 & 71.85 & 67.11 &  3.46 \\
\midrule
\rowcolor{gray!50}
Mean $\Delta$ (\%)   & +1.13 & +1.20 & +14.81 & +2.90 & +3.80 & +13.64 & +2.58 & +1.93 & +0.83 \\
\rowcolor{gray!50}
Min $\Delta$ (\%)   & +0.64 & +0.60 & +1.92  & +2.42 & +2.38 & +5.20  & +1.44 & +0.75 & +0.46 \\

\bottomrule
\end{tabular}
\label{tab:sd_filtered}
\vspace{-6pt}
\end{table*}

\section{Experiments} 
\label{sec:experiments}

\subsection{Setup} \label{exp:setup}

We tested our findings on \textbf{six datasets} spanning image resolutions from 
$32^2$ to $512^2$ 
(see Appendix~\ref{sec:datasets and splits} for detailed dataset descriptions and train/validation splits).
All experiments are conducted under four distinct threat models and executed on eight NVIDIA~A40 GPUs.

\paragraph{Evaluation Metrics: }We evaluated attack performance using several metrics: ASR (attack success rate, i.e., membership inference accuracy), AUC (Area Under ROC Curve), TPR at 1\% FPR (TPR@1\%FPR), and the number of queries per attack (\#Query).  The TPR@1\%FPR is computed by selecting the threshold~$\tau$ at which the false positive rate falls just below $0.01$, and reporting the corresponding true positive rate at that operating point. 
The ASR is defined as the maximum accuracy achieved over all thresholds, i.e.\ $\tau^* = \max_{\tau} \tfrac{1}{2}\big(\text{TPR}(\tau) + 1 - \text{FPR}(\tau)\big)$ in the balanced setting. The AUC is computed as the trapezoidal integral of $\text{TPR}(\tau)$ against $\text{FPR}(\tau)$ across all thresholds.

\paragraph{Latent Diffusion Model:}
For \mbox{CIFAR‑10} and CelebA \citep{liu2015deep}, we trained a $\beta$-VAE \cite{higgins2017beta} from scratch on the \textit{member} set. Then we use the latent representations of member set to train a latent space diffusion model unconditionally while VAE encoder-decoder is frozen. From each training split we subsample $n$ images and partition them equally into a \textit{member} set and a \textit{held‑out} set. ImageNet-1K \citep{russakovsky2015imagenet} is trained at two different points: (1) We directly employ the publicly released autoencoder model \texttt{stabilityai/sd-vae-ft-mse}~\cite{stabilityai_sdvae_ft_mse} without further fine-tuning. This model, developed by Stability~AI, serves as the standard latent autoencoder used in the Stable Diffusion~v1 framework. (2) The latent space diffusion model is also trained from scratch. We train it conditonally by using the architecture components from diffuser, which we guarantee have a similar architecture to vanilla LDM \cite{rombach2022high}. The KL-regularization ($\beta$) for three datasets are $10^{-2}$, $10^{-3}$, $10^{-6}$. With higher resolution, one expects to reduce the $\beta$ to ensure the same scale of regularization across datasets. and The architecture details are released in Appendix \ref{sec:arch_details_ldm}. We also implement the proposed method on LDMs with VQ-GAN~\cite{esser2021taming} as the encoder (See details in Appendix~\ref{sec:ldm_vqgan}).

\paragraph{Pre-trained Latent Diffusion Models:}
For Pokémon\cite{huggingface_pokemon}, COCO2017‑Val \citep{lin2014microsoft}, and Flickr30k \citep{young2014image}, we fine‑tuned Stable Diffusion v1‑4\footnote{\url{https://huggingface.co/CompVis/stable-diffusion-v1-4}} on a randomly selected subset of each training split, reserving an equally sized subset as the held‑out set. None of these datasets are in the original pre‑training corpus. 

Stable Diffusion v1‑4 was fine‑tuned for \textbf{15k}, \textbf{100k}, and \textbf{200k} steps on the member splits of Pokémon, COCO‑2017‑Val, and Flickr30k, respectively, with a fixed learning rate of $10^{-5}$ (AdamW). Additionally, we adopted the default data augmentation (Random-Crop and Random-Flip) while training.

\subsection{Experiments on distortion and memorization} \label{exp:local_dist}
To compute the local distortion at each latent point, we employ our matrix-free randomized SVD procedure (Appendix~\ref{alg:rand-svd-jacobian}). This algorithm evaluates Jacobian-vector products using \texttt{jvp}/\texttt{vjp} primitives in PyTorch, thereby avoiding explicit construction of the full decoder Jacobian. When automatic differentiation is unavailable, we find that finite-difference approximations provide a sufficiently accurate substitute for \texttt{jvp}. Unless otherwise stated, we use fixed randomized SVD hyperparameters across all experiments: target rank $k=20$, oversampling parameter $p=30$, and $q=2$ power iterations, which balance computational efficiency and numerical stability. For measuring memorization, we primarily adopt SimA~\cite{rao2025score} due to its simplicity and effectiveness.

Figure~\ref{fig:mia_distort} summarizes our results. For each dataset, we perform membership inference attacks at multiple diffusion timesteps $t$ (x-axis) and report the resulting AUC (y-axis). We partition both the member set and the held-out set into four quartiles according to their local distortion values and evaluate attacks on each quartile independently. We set the quartile threshold through a joint set of member and held-out. For example, the top (75\%--100\%) quartile corresponds to samples whose local distortion lies in the top quartile across all available points; the AUC reported for this group is computed using only the members and held-out samples that fall into this distortion range. As a baseline, we construct a ``random'' group by repeatedly (ten times) sampling equal numbers of members and non-members and reporting the mean AUC with one standard deviation (shown as a shaded band).

Across all datasets, we observe a consistent and pronounced ordering: higher-distortion quartiles yield substantially higher attack AUCs, indicating that regions of large decoder distortion exhibit stronger memorization. The magnitude of this effect varies by dataset. For CIFAR-10, the distortion gap between quartiles is large (Fig.~\ref{fig:distortion}), producing correspondingly large separations in attack performance; in contrast, CelebA exhibits nearly uniform distortion throughout its latent space, resulting in only minor differences in AUC across quartiles. We leave MNIST high/low distortion samples in the appendix. Interestingly, the lines on ImageNet-1K experiments cross at one point. This weird crossing is dataset-dependent. Given that the earlier time MIA is not trustworthy~\cite{rao2025score}, it still shows a consistent phenomenon after passing the early time. This strong correlation between decoder's local distortion and memorization indicates that the choice of encoder--decoder architecture plays a central role in shaping the learned score field and, consequently, the memorization behavior of latent diffusion models.



\subsection{Experiments on per-dimension memorization} \label{exp:per_dim_dist}

\paragraph{Benchmark methods: } 
To justify our work, we selectively choose four state-of-the-art threat models: SimA \cite{rao2025score}, SecMI \cite{duan2023diffusion}, PIA \cite{kong2023efficient}, and Loss (also refered to ``Naive'' method) \citep{matsumoto2023membership} as our benchmark methods
acknowledging that list is not exhaustive. For each method, we followed the hyperparameter suggestion in their original paper. Notably, $l_4\text{-norm}$, $l_2\text{-norm}$, $l_4\text{-norm}$ and $l_2\text{-norm}$ were used for SimA, SecMI, PIA and Loss as suggested.

The complete results are reported in Tables~\ref{tab:ldm_filtered} and~\ref{tab:sd_filtered}. We report the best performance over probe time $t=[0:300]$. Throughout all experiments, we apply a dilution procedure, which removes the top-$k\%$ least influential latent dimensions from the attack vector prior to computing the norm. Although we do not conduct a search for the optimal value of $k$, we expect that such an optimal choice may vary across datasets and individual samples; exploring this dependency is an interesting direction for future work. For simplicity and consistency, we thus adopt $k=40$ as a global setting. As a control, we also include a baseline that randomly discards $40\%$ of dimensions when applying SimA (first row in each table), which typically reduces attack performance (A table of various filtering ratios $k$ is provided in the Appendix~\ref{sec:ablation_k}). In contrast, removing the least-memorizing dimensions based on our proposed per-dimension metric consistently improves all three evaluation measures across all methods and datasets. And from the ``min $\delta$'', we found that Loss method usually achives the least improvement compared to others. These results indicate that latent-space diffusion models exhibit dimension-wise non-uniform memorization, with certain latent coordinates contributing disproportionately to overfitting and membership leakage.

\subsection{Connection to high-frequency components} \label{exp:high_freq}

\citet{lian2025unveiling} reported that reconstructing images from denoised noise vectors and suppressing high-frequency components made members and non-members more separable. This approach performed well for diffusion models trained directly in the data domain, but in our experiments it reduced performance for more complex latent-space diffusion models (e.g., Stable Diffusion); see Section \ref{sec:attack_suppress_high_freq} in the appendix. We hypothesize that if the latent encoder were perfectly linear, then manipulating Fourier components would have behaved as expected in latent space. However, the non-linear VAE encoder used in latent diffusion models does not simply map images to Fourier modes; instead, it performs denoising on a sub-manifold of the data domain for which the spectral analog is generally inaccessible (the spectrum of the Laplace-Beltrami operator).

To test this hypothesis, we computed the correlation between the local distortion of each image and the summed energy of its low-frequency and high-frequency spectra, for each of the three datasets CIFAR-10, CelebA, and ImageNet-1K. To select low frequencies, we summed the center elements of the Fourier transformed data; for CIFAR-10 ($32^2$), CelebA ($64^2$) and ImageNet-1K ($256^2$) the central radii summed were 5, 10, 40, which are approximately $2.5\%$ of the spectra for each.
The results, summarized in Table~\ref{tab:correlation_dist_hf}, show that at a high level the distortion reflects the intuition of \citet{lian2025unveiling}, but that this analog is imperfect, especially for more structured datasets such as CelebA, where all data have aligned structure (spatially registered faces, in the case of CelebA).

\begin{table}[!t]
\centering
\caption{Pearson correlation coefficients ($r$) between the decoder distortion
and high/low-frequency amplitudes across datasets. The chosen threshold radius between low and high frequency for CIFAR-10 ($32^2$), CelebA ($64^2$) and ImageNet-1K ($256^2$) are 5, 10, 40, which are approximately $2.5\%$ of the spectra for each. }
\setlength{\tabcolsep}{6pt}
\renewcommand{\arraystretch}{1.05}
\scriptsize
\begin{tabular}{lcc}
\toprule
Dataset & $r(\text{Distortion}, \text{LF})$ & $r(\text{Distortion}, \text{HF})$ \\
\midrule
CIFAR-10    & 0.0814 & 0.7156 \\
CelebA      & -0.8713 & -0.2865 \\
ImageNet    & 0.1501 & 0.6440 \\
\bottomrule
\end{tabular}
\label{tab:correlation_dist_hf}
\vspace{-4pt}
\end{table}

\section{Discussion} 
\label{sec:disc}

Our results suggest that membership inference and thereby model inversion for latent diffusion models may directly connected to the construction of the latent space itself.
This runs counter to current thinking in attack methods, which apply the same methods to data-domain diffusion (e.g., the original DDPM model) and latent diffusion models.
Instead, these results suggest that an optimal attack method would account for decoder geometry in its approach.

Distortion filtering uniformly improved performance across methods. This is surprising, not only because disparate approaches can be improved by the same influence measure filter, but also because the decoder from which the pullback metric distortion is computed is \emph{not directly involved in the diffusion model fitting} itself. Common practice prescribes fitting a VAE to the dataset and then freezing the encoder/decoder pair, meaning the diffusion model fitting and any associated overfitting occurs on the latent space as generic data points; the diffusion model is not aware that it is operating on a latent code instead of data-domain points.

The necessity of accounting for decoder geometry for LDM inversion is therefore due to changes in distributional structures $p(z)$ versus $p(x)$, not diffusion dynamics. In the unlikely cases where a dataset had the same distribution as these VAEs, or an ill-fit VAE had the same distributional structure as the data domain, their diffusion models would have correspondingly matching dynamics. Because these cases are unlikely, in practice we conclude that inversion should account for decoder geometry, and more broadly that the overfitting phenomena in latent diffusion models is connected to the latent space.


\textbf{Limitations and Caveats:} We have tested only a subset of attack methods and a subset of possible datasets. Conditionalization, dataset size, and dataset structure clearly influence diffusion model behavior and thereby inversion dynamics. We can clearly design datasets where the data domain is similar to these latent spaces, and we can design trivial ``VAEs'' with no curvature. Our results are likely relevant for similar image data, which represent the mainstay of diffusion model applications, but they may not generalize to disparate domains (e.g., discrete diffusion on text or tabular data).

\textbf{Crossover strata:} The ImageNet subfigure of Figure \ref{fig:mia_distort} (at far right) show that the distortion strata lines cross over each other around timestep 70. This seems to inform upon interactions between the noise schedule and the length-scale of the latent space; clearly for faster noise schedules this cross-over point would occur earlier. Interactions between noise schedules and decoder geometry may be important for attack parameter tuning and overall understanding of overfit phenomena.

\section{Conclusion} 
\label{sec:conclusion}

We show that latent diffusion models exhibit spatially non-uniform memorization in latent space and demonstrate that this behavior is tightly correlated with the local distortion measured by the decoder pullback metric.
We propose a principled, geometry-driven measure of per-dimensional memorization strength based on each latent dimension's contribution to decoder distortion.
We provide extensive empirical evidence showing that removing less-memorizing latent dimensions of attack vector before computing the norm statistics consistently improves all metrics over all MIA methods.

These findings directly inform upon membership inference and model inversion in latent diffusion models, demonstrating the importance of the latent space and the decoder geometry to these problems.

\section{Acknowledgements}

This work was supported in part by NSF 2321684 and the ARPA-H ALISS project.

{
    \small
    \bibliographystyle{ieeenat_fullname}
    \bibliography{main}
}

\clearpage
\setcounter{page}{1}
\maketitlesupplementary

\renewcommand\thesection{\Alph{section}}
\setcounter{section}{0}

\section{Approximation of Local Distortion}
\label{sec:approx_local_distor}
An pseudo-code to efficiently compute local distortion using randomized singular value decomposition is provided in Algo. \ref{alg:rand-svd-jacobian}. Noticeably, this code computing Jacobian–vector products using jvp/vjp packages in pytoch without explicitly computing the Jacobian matrix. In our experiments, finite-difference is also a good approximation of jvp when automatic differentiation is unavailable

\begin{algorithm}*[t]
\caption{Matrix–Free Randomized SVD for Top-$k$ Singular Values of $J_D(z)$}
\label{alg:rand-svd-jacobian}
\small
\begin{algorithmic}[1]
\Require Decoder mean map $D:\mathcal{Z}\!\to\!\mathcal{X}$; latent point $z\!\in\!\mathcal{Z}$; target rank $k$; oversampling $p$; power passes $q$; flag \textsc{use\_jvp}$\in\{\textsc{true},\textsc{false}\}$; finite-difference step $h$
\Ensure Approx.\ top-$k$ singular values $\{s_i\}_{i=1}^k$ of $J_D(z)$ and $\log$–volume $\sum_{i=1}^k \log s_i$
\State $d \gets \dim(\mathcal{Z}),\quad m \gets \dim(\mathcal{X}),\quad \ell \gets \min(k{+}p,\, d)$
\State \textbf{define} decoder–Jacobian oracles at $z$:
\Statex \hspace{1.25em}$\mathcal{J}(V)=J_D(z)V \in \mathbb{R}^{m\times r}$ \textbf{via} JVP if \textsc{use\_jvp}=\textsc{true}, \textbf{else} central differences: $[(D(z{+}h v_j)-D(z{-}h v_j))/(2h)]_j$
\Statex \hspace{1.25em}$\mathcal{J}^\top(Y)=J_D(z)^\top Y \in \mathbb{R}^{d\times r}$ \textbf{via} VJP (reverse-mode autodiff)
\State Draw $V_0 \sim \mathcal{N}(0,I_{d\times \ell})$; \quad $V \gets \mathrm{qr}(V_0)$ \Comment{$V\in\mathbb{R}^{d\times \ell}$}
\For{$t=1$ to $q$} \Comment{optional power iteration on $G=J_D^\top J_D$}
    \State $Y \gets \mathcal{J}(V)\in\mathbb{R}^{m\times \ell}$;\quad $G \gets \mathcal{J}^\top(Y)\in\mathbb{R}^{d\times \ell}$
    \State $V \gets \mathrm{qr}(G)$
\EndFor
\State $Y \gets \mathcal{J}(V)\in\mathbb{R}^{m\times \ell}$;\quad $Q \gets \mathrm{qr}(Y)\in\mathbb{R}^{m\times \ell}$
\State $T \gets \mathcal{J}^\top(Q)\in\mathbb{R}^{d\times \ell}$
\State $\{\hat{s}_i\}_{i=1}^{\ell} \gets \mathrm{svdvals}(T)$; \textbf{sort} $\hat{s}_i$ in descending order
\State $s_i \gets \hat{s}_i$ for $i=1,\dots,k$ \Comment{top-$k$ singular values of $J_D(z)$}
\State \textbf{return} $\{s_i\}_{i=1}^k$, \quad $\log$–volume $= \sum_{i=1}^{k} \log s_i$
\end{algorithmic}
\end{algorithm}

\section{Variation of Filtering Ratio $k$}
\label{sec:ablation_k}
We evaluated masking ratios $k \in [10\%, 90\%]$ on ImageNet-1K(subset) of SimA~\cite{rao2025score} attack statistics. The results are summarized on Table~\ref{tab:masking_ratios}

\begin{table*}[t]
\centering
\caption{AUC performance of SimA attack under different masking ratios $k$ on ImageNet-1K (subset). For comparison, the last column presents a reverse-masking setting where the top 90\% most-memorizing dimensions are removed. Retaining only the bottom 10\% renders the attacks almost ineffective.}
\label{tab:masking_ratios}
\begin{tabular}{@{}lcccccccccc|c@{}}
\toprule
Masking Ratio $k$        & 0.0   & 0.1   & 0.2   & 0.3   & 0.4   & 0.5   & 0.6   & 0.7   & 0.8 & 0.9 & \textbf{0.9 (Rev)} \\ \midrule
AUC ($\uparrow$)         & 69.70 & 70.77 & 71.68 & 72.32 & 72.64 & 72.54 & 71.95 & 70.65 & 68.91 & 65.99 & \textbf{54.67} \\
TPR@1\%FPR ($\uparrow$)  & 3.96  & 4.68  & 5.64  & 7.06  & 7.79  & 8.09  & 8.62  & 9.04  & 8.46  & 6.78  & \textbf{2.21}  \\ \bottomrule
\end{tabular}
\end{table*}

\section{Datasests and splits}
\label{sec:datasets and splits}
The dataset and member/non-member splits information are provided in Table~\ref{tab:dataset_splits}.

\begin{table*}[t]  
\centering
\caption{Datasets and splits used for our experiments. \textbf{VAE} denotes the \textbf{VAE} of LDMs. \textbf{LM} denotes \textbf{Diffusion Part of LDMs}}
\label{tab:dataset_splits}
\small
\resizebox{\linewidth}{!}{%
\begin{tabular}{l
                c c
                l l
                l
                c c}
\toprule
\textbf{Model}   &
\textbf{Member}  & \textbf{Held-out} &
\textbf{Pre-trained (VAE)} & \textbf{Pre-trained(LM)} &
\textbf{Splits}      &
\textbf{Resolution}  & \textbf{Cond.}  \\
\midrule
\multirow{4}{*}{Latent Diffusion Models} %
  & MNIST & MNIST & No       & No          & 30k/30k & 28  & --   \\
  & CIFAR-10 & CIFAR-10 & No       & No          & 25k/25k & 32  & --   \\
  & CelebA & CelebA & No  & No          & 30k/30k & 64  & --   \\
    & ImageNet-1k(train) & ImageNet-1K(Val) & Yes & No & 100k/100k & 256 & class \\
\midrule
\multirow{2}{*}{Stable Diffusion V1-4} %
  &   Pokémon  &   Pokémon  & Yes & Yes      & 416/417  & 512 & text \\
  & COCO2017-Val & COCO2017-Val & Yes & Yes & 2.5k/2.5k & 512 & text \\
  & Flickr30k & Flickr30k & Yes & Yes & 10k/10k & 512 & text \\
\bottomrule
\end{tabular}}
\end{table*}

\section{Per–Dimension Influence via Hutchinson Estimator}
\label{sec:hutchinson_est}
A pseudocode of our method to compute the per-dimension influence on memorization via the Hutchinson Estimator is provided in Algo. \ref{alg:perdim-hutch}.

\begin{algorithm}[t]
\caption{Per–Dimension Influence via Hutchinson Estimator}
\label{alg:perdim-hutch}
\small
\begin{algorithmic}[1]
\Require Decoder $D$, latent samples $\{z^{(n)}\}_{n=1}^{N}$ with $z^{(n)}\!\in\!\mathbb{R}^{d}$, number of Monte Carlo probes $n_{\mathrm{mc}}$, stability constant $\varepsilon>0$
\Ensure Matrix $\mathrm{PerDim}\!\in\!\mathbb{R}^{N\times d}$ containing per–dimension log–influence for each $z^{(n)}$
\For{$n = 1$ to $N$} \Comment{Process each latent code independently}
  \State $z \gets z^{(n)}$; enable gradient on $z$
  \State $x \gets D(z)$; $\quad x \in \mathbb{R}^{m}$ \Comment{Decoder output (flattened)}
  \State $\mathrm{diag\_est} \gets \mathbf{0}\in\mathbb{R}^{d}$ \Comment{Accumulator for $\operatorname{diag}(J_D^\top J_D)$}
  \For{$j = 1$ to $n_{\mathrm{mc}}$} \Comment{Hutchinson probes}
    \State $v \sim \mathcal{N}(\mathbf{0}, I_m)$ \Comment{Random probe in output space}
    \State $g \gets J_D(z)^{\top} v$ \Comment{Compute VJP via reverse–mode autodiff}
    \State $\mathrm{diag\_est} \gets \mathrm{diag\_est} + g \odot g$ \Comment{Elementwise square and accumulate}
  \EndFor
  \State $\mathrm{diag\_est} \gets \mathrm{diag\_est} / n_{\mathrm{mc}}$ \Comment{Monte Carlo average}
  \State $\mathrm{per\_dim\_log} \gets \tfrac{1}{2}\log(\mathrm{diag\_est} + \varepsilon)$ \Comment{Log–compressed influence}
  \State $\mathrm{PerDim}[n,:] \gets \mathrm{per\_dim\_log}$
\EndFor
\State \Return $\mathrm{PerDim}$
\end{algorithmic}
\end{algorithm}

\section{Experiments on LDMs with VQ-GAN as the encoder}
\label{sec:ldm_vqgan}
For experiments on MNIST, CIFAR-10 and CelebA, we used a vanilla $\beta$-VAE. For experiments on ImageNet, Pokémon, COCO‑2017‑Val, and Flickr30k, we employed Stability AI's autoencoders trained with \textbf{Adversarial (GAN) and Perceptual (LPIPS) losses}. We also test our masking strategy on VQ-GAN~\cite{esser2021taming}(discrete, large geometry change). The original LDM implementation used the bottleneck layer $\hat{z}$ before quantization; since we need the decoder side as well, we approximate $\frac{\partial D(\hat{z})}{\partial \hat{z}} \approx \frac{\partial D(z_q)}{\partial z_q}$ after quantization. We trained a VQ-GAN from scratch for CIFAR-10 and MNIST and used a pre-trained one\footnote{\url{https://ommer-lab.com/files/latent-diffusion/vq-f8.zip}} for ImageNet-1K. Training details, such as the number of training steps, are kept consistent with the other experiments. Table~\ref{tab:vq_gan} shows that our method consistently looks for the most-memorizing dimension, which corresponds to the per-dimensional distortion.

\begin{table}[t] 
\centering
\caption{\textbf{LDMs with VQGAN as the encoder}. Performance (AUC \%) comparison on different datasets. Numbers in bold indicate improvement.}
\label{tab:vq_gan}
\resizebox{\columnwidth}{!}{
\begin{tabular}{lccc}
\toprule
Method & MNIST & CIFAR-10 & ImageNet-1K \\
\midrule
SimA (After Filter) & 83.69 (\textbf{+2.7}) & 86.27 (\textbf{+3.2}) & 67.38 (\textbf{+3.85}) \\
\bottomrule
\end{tabular}
}
\end{table}

\section{Latent diffusion attack by smoothing high-frequency components}
\label{sec:attack_suppress_high_freq}
In this section, we apply the method proposed by \citet{lian2025unveiling} to latent diffusion models (LDMs), specifically Stable Diffusion. Although the method was originally developed for data-domain diffusion models, we extend it to the latent domain using the following formulation.

\noindent \textbf{Loss:}
The attack statistics can be expressed as, for $\varepsilon\sim\mathcal N(0,I)$:
\begin{equation}
\text{Loss}=
    \|\epsilon - \epsilon_\theta(\sqrt{\bar{\alpha}_t} z_0 + \sqrt{1-\bar{\alpha}_t}\,\epsilon,\, t) \|,
\end{equation}
where $\epsilon_\theta(\cdot)$ indicates the predicted noise. According to Appendix B of \citet{lian2025unveiling}. We derive

\begin{equation}
    z_0^{\text{target}}
    = \frac{z_t - \sqrt{1-\bar{\alpha}_t}\,\epsilon}{\sqrt{\bar{\alpha}_t}},
\end{equation}
\begin{equation}
    z_0
    = \frac{z_t - \sqrt{1-\bar{\alpha}_t}\,\epsilon_\theta
        ( \sqrt{\bar{\alpha}_t} z_0 + \sqrt{1-\bar{\alpha}_t}\,\epsilon,\, t )}
        {\sqrt{\bar{\alpha}_t}}.
\end{equation}

To smooth the high-frequency components, we should map latent codes $z$ to the image $x$. Then, $x_0=D(z_0) \text{ and }x_0^{\text{target}}=D(z_0^{\text{target}})$.

\noindent \textbf{PIA:} Similarly, it can be expressed as:
\begin{equation}
    \text{PIA}=
    \left\|
        \epsilon_\theta(z_0, 0)
        - \epsilon_\theta\!\left( 
            \sqrt{\bar{\alpha}_t}\, z_0 
            + \sqrt{1-\bar{\alpha}_t}\,\epsilon_\theta(z_0,0),\,
            t
        \right)
    \right\|,
\end{equation}
where $\epsilon_\theta(z_0,0)$ represents the noise prediction for $z_0$. According to Appendix B of \citet{lian2025unveiling}, it can be converted to
\begin{equation}
    z_0^{\text{target}}
    = \frac{z_t - \sqrt{1-\bar{\alpha}_t}\,\epsilon_\theta(z_0,0)}{\sqrt{\bar{\alpha}_t}},
\end{equation}

\begin{equation}
    z_0
    = \frac{
        z_t
        - \sqrt{1-\bar{\alpha}_t}\,\epsilon_\theta\!(
            \sqrt{\bar{\alpha}_t} z_0
            + \sqrt{1-\bar{\alpha}_t}\,\epsilon_\theta(z_0,0),\,
            t
        )
    }{
        \sqrt{\bar{\alpha}_t}
    }.
\end{equation}
Likewise, $x_0=D(z_0) \text{ and }x_0^{\text{target}}=D(z_0^{\text{target}})$.

A high-frequncy filter is defined the same as \citet{li2024unveiling},
\begin{equation}
    \mathcal{F}(x) = \mathrm{IFFT}\!\left( \mathrm{FFT}(x) \odot \beta(r) \right),
\end{equation}
where FFT and IFFT denotes the \textit{Fast Fourier Transform} and \textit{Inverse Fast Fourier Transform}. $\odot$ denotes element-wise multiplication, and $\beta_{i,t}(r)$ is a mask that keeps the low-frequency components and smooths the high-frequency components. Specifically,
\begin{equation}
    \beta(r) = 
    \begin{cases}
        s, & \text{if } r > r_t, \\
        1, & \text{otherwise}.
    \end{cases}
\end{equation}
where $s$ is frequency-dependent scaling factor, $r$ denotes the radius of frequency-domain, 
and $r_t$ is the high-frequency threshold radius. Then the attack statistics for both Loss and PIA methods become:
\begin{equation}
    \| \mathcal{F}(x_0) - \mathcal{F}(x_0^{\text{target}}) \|.
\end{equation}

Under the assumption that latent-space diffusion models inherit the same learning dynamics as data-domain diffusion models, one would expect the proposed high-frequency filtering technique to be equally effective in both settings. Contrary to this expectation, our experiments show that the method decreases the attack performance for Stable Diffusion (Table~\ref{tab:sd_high_freq_smooth}). This divergence suggests that the mapping from the data domain to the latent domain—typically implemented through a nonlinear VAE encoder—introduces non-trivial distortions that alter the frequency structure relevant for membership inference. Consequently, careful consideration of this domain transformation is essential when adapting techniques developed for data-domain models.



\begin{table*}[!t]
\centering
\caption{Attack performance of four baseline methods used to evaluate memorization of \textbf{Stable Diffusion}.}
\setlength{\tabcolsep}{5pt}
\renewcommand{\arraystretch}{1.05}
\scriptsize
\begin{tabular}{lccc|ccc|ccc}
\toprule
\multirow{2}{*}{Method} &
\multicolumn{3}{c|}{Pokémon (\%)} &
\multicolumn{3}{c|}{MS-COCO (\%)} &
\multicolumn{3}{c}{Flickr (\%)} \\
\cmidrule(lr){2-4}\cmidrule(lr){5-7}\cmidrule(lr){8-10}
& AUC↑ & ASR↑ & TPR@1\%FPR↑
& AUC↑ & ASR↑ & TPR@1\%FPR↑
& AUC↑ & ASR↑ & TPR@1\%FPR↑ \\
\midrule

Loss  & 92.03 & 85.47 & 40.77
      & 83.26 & 75.80 & 13.48
      & 63.13 & 59.74 & 2.35 \\
\rowcolor{gray!20}
Loss (+high freq. filter)  & 84.39 & 78.40 & 19.90
      & 73.46 & 68.08 & 4.32
      & 57.52 & 55.81 & 1.61 \\
PIA   & 94.93 & 90.52 & 32.85
      & 92.51 & 85.72 & 24.76
      & 68.88 & 64.61 & 2.67 \\
\rowcolor{gray!20}
PIA (+high freq. filter)  & 89.81 & 84.27 & 19.90
      & 73.32 & 67.72 & 4.96
      & 57.37 & 55.61 & 1.43 \\

\bottomrule
\end{tabular}
\label{tab:sd_high_freq_smooth}
\vspace{-6pt}
\end{table*}

\section{More image examples on different local distortion}
\label{sec:examples_local_distorch}

\begin{figure*}[t]
    \centering
    \includegraphics[width=\textwidth]{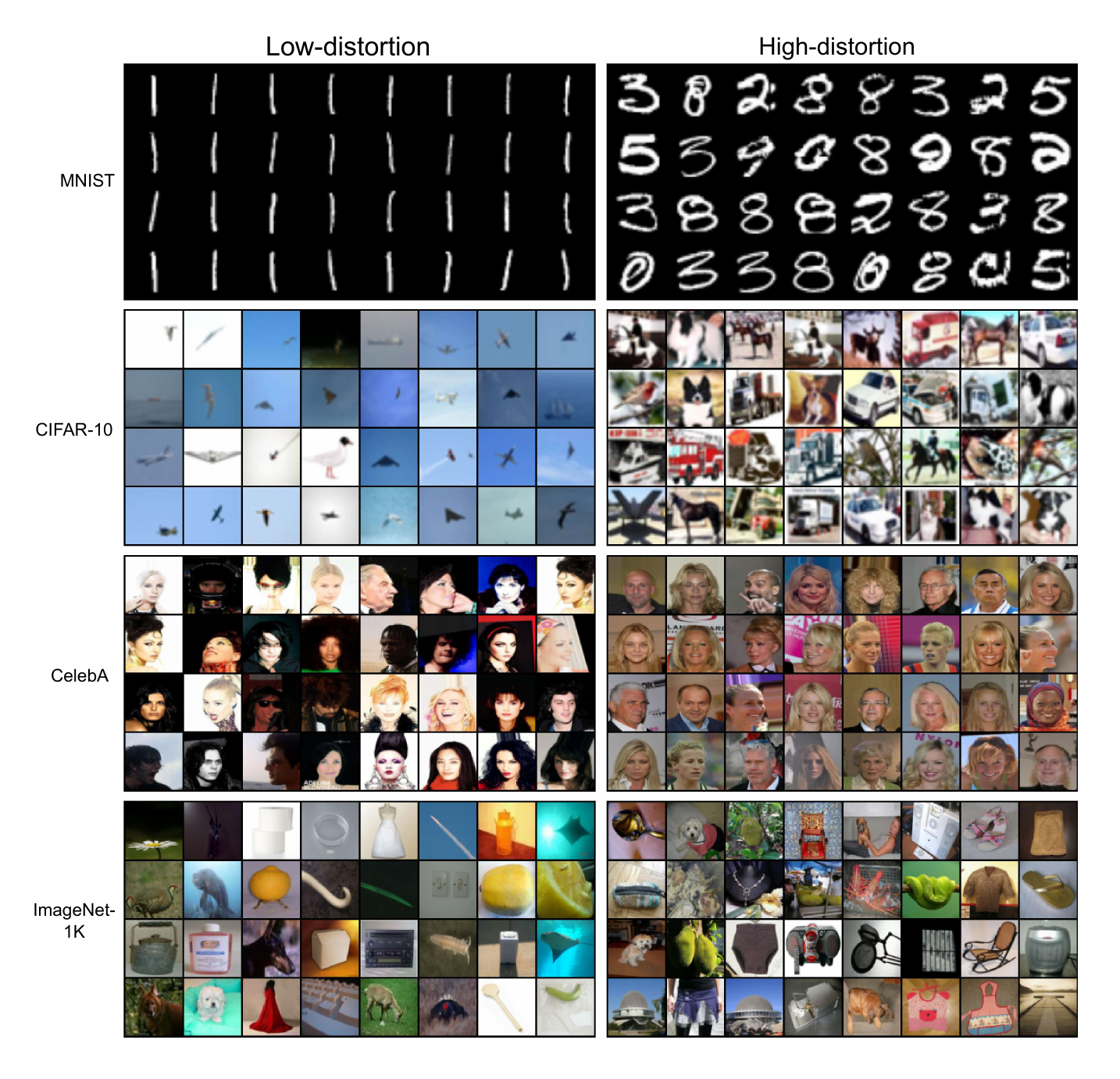}
    \caption{More examples of images in high (low) distortion region}

    \label{fig:more_ex_distortion}
\end{figure*}

More examples of images in the high (low) distortion region are provided in Fig. ~\ref{fig:more_ex_distortion}. In a standard VAE architecture, the latent space is a bottleneck, meaning $k < d$. If $x = D(z)$ is purely deterministic, the generated data does not span the full $\mathbb{R}^d$ space. Instead, it lies strictly on a $k$-dimensional manifold embedded within the $d$-dimensional ambient space.

Because of this, $p(x)$ is a degenerate distribution in $\mathbb{R}^d$ (it has zero volume) and the standard change of variables formula fails. However, if we evaluate the density strictly on the manifold with respect to the $k$-dimensional Hausdorff measure, we must use the area formula. The metric tensor induced on the latent space by the mapping $D$ is $J_D(z)^T J_D(z)$.The density on the manifold is~\cite{brehmer2020flows, graham2017asymptotically}:
$$
p(x) = p(z) \left( \det \left( J_D(z)^T J_D(z) \right) \right)^{-1/2}
$$
Taking the logarithm gives:
\begin{equation}
\log p(x) = \log p(z) - \frac{1}{2} \log \underbrace{\det \left( J_D(z)^T J_D(z) \right)}_{\text{Distortion(z)}}
\label{eq:vae_log_likelihood}
\end{equation}
According to \cite{rao2026generalization, nalisnick2018deep, serra2019input}, the log-likelihood estimation in a generative model is highly biased toward the input complexity, where the model assigns more density to complex samples. Based on Eq.~\ref{eq:vae_log_likelihood}, this insight explains the positive correlation between visual complexity and the distortion of a sample in Fig.~\ref{fig:distortion} and Fig.~\ref{fig:more_ex_distortion}. Hence, the distribution shifts significantly from the data space to the latent space.

\section{Architecture details of all LDM instances}
\label{sec:arch_details_ldm}

\paragraph{CIFAR-10.}
We train a convolutional VAE at $32\times 32$ RGB resolution with 3 input channels, base width 128, and a 4-channel latent space. The encoder consists of strided convolutions and residual blocks with GroupNorm and SiLU activations, downsampling $32\!\to\!16\!\to\!8$; the decoder mirrors this structure. The VAE is optimized for 120 epochs on the member subset using AdamW (learning rate $2\times 10^{-4}$, $\beta=(0.9, 0.999)$, weight decay $10^{-4}$), with an $\ell_1$ reconstruction term and a KL divergence term weighted by $\beta_{\mathrm{KL}} = 10^{-2}$. On the frozen latent space, we train a DDPM-style UNet with 128 base channels, channel multipliers $(1,2,2,2)$, two residual blocks per resolution level, dropout 0.1, and GroupNorm. The diffusion model uses a linear noise schedule with $T=1000$ steps and $\beta \in [10^{-4}, 2\times 10^{-2}]$, and is trained with AdamW (same hyperparameters) for 2048 epochs with batch size 128.

\paragraph{CelebA.}
For CelebA, we first center-crop faces to $140\times 140$, resize to $64\times 64$, and apply random horizontal flips and per-channel normalization to $[-1,1]$. We train a VAE with the same architectural configuration (3 input channels, base width 128, 4-channel latent space) for 120 epochs using AdamW (learning rate $2\times 10^{-4}$, $\beta=(0.9, 0.999)$, weight decay $10^{-4}$), but with a smaller KL weight $\beta_{\mathrm{KL}} = 10^{-3}$. The latent diffusion model is a UNet with 224 base channels and channel multipliers $(1,2,3,4)$, two residual blocks per level, dropout 0.1, and a linear noise schedule with $T=1000$ diffusion steps and $\beta \in [10^{-4}, 2\times 10^{-2}]$. It is trained with AdamW (same optimizer settings) for 512 epochs with batch size 256.

\paragraph{ImageNet-1K.}
We train a class-conditional latent diffusion model on ImageNet-1K using the pretrained Stable Diffusion VAE (\texttt{sd-vae-ft-mse}) as a frozen encoder--decoder (with default $\beta_{KL}=1e-6$). Input images are resized and center-cropped to $256\times 256$ RGB, augmented with random horizontal flips, and normalized to $[-1,1]$. The VAE deterministically encodes images into $4\times 32\times 32$ latents (8$\times$ spatial downsampling), which are scaled by $0.18215$ before diffusion and rescaled back when decoding.

On this latent space, we train a UNet with 4 input/output channels, 2 residual blocks per resolution, and block widths $(192, 384, 576, 960)$, using cross-attention layers with a learned class embedding of dimension 512. Class conditioning is implemented via an embedding table over the 1{,}000 ImageNet labels, passed as the UNet cross-attention context. The diffusion process uses $T=1000$ steps with a linear noise schedule $\beta \in [1.5\times 10^{-3}, 1.95\times 10^{-2}]$.

We construct a balanced training subset with 50 images per class (50k images total) from the ImageNet-1K training split and use the standard validation split for evaluation and qualitative sampling. The UNet is optimized for 600 epochs with AdamW (learning rate $10^{-4}$, $\beta=(0.9, 0.999)$, zero weight decay) and a cosine learning-rate schedule with 1{,}000 warm-up steps. Training is performed with mixed-precision (FP16) and distributed data parallelism with an effective per-device batch size of 21.

\paragraph{Fine-tuning on Stable diffusion.} We reuse the Stable Diffusion V1-4 architecture for all stable diffusion experiments.


\end{document}